\documentclass{bmvc2k}

\usepackage{multirow}
\usepackage{amsmath}
\usepackage{amssymb}
\usepackage{amsfonts}

\title{MSMVD: Exploiting Multi-scale Image Features via Multi-scale BEV Features for Multi-view Pedestrian Detection}

\addauthor{Taiga Yamane}{taiga.yamane@ntt.com}{1}
\addauthor{Satoshi Suzuki}{satoshixv.suzuki@ntt.com}{1}
\addauthor{Ryo Masumura}{ryo.masumura@ntt.com}{1}
\addauthor{Shota Orihashi}{shota.orihashi@ntt.com}{1}
\addauthor{Tomohiro Tanaka}{tomohiro.tanaka@ntt.com}{1}
\addauthor{Mana Ihori}{mana.ihori@ntt.com}{1}
\addauthor{Naoki Makishima}{naoki.makishima@ntt.com}{1}
\addauthor{Naotaka Kawata}{naotaka.kawata@ntt.com}{1}

\addinstitution{
 NTT Human Informatics Laboratories\\
 NTT Corporation\\
 Yokosuka, Japan
}

\runninghead{Yamane et al.}{MSMVD}

\def\eg{\emph{e.g}\bmvaOneDot}

\def\etal{\emph{et al}\bmvaOneDot}

\begin{document}

\maketitle

\begin{abstract}
Multi-View Pedestrian Detection (MVPD) aims to detect pedestrians in the form of a bird's eye view (BEV) from multi-view images.
In MVPD, end-to-end trainable deep learning methods have progressed greatly.
However, they often struggle to detect pedestrians with consistently small or large scales in views or with vastly different scales between views.
This is because they do not exploit multi-scale image features to generate the BEV feature and detect pedestrians.
To overcome this problem, we propose a novel MVPD method, called Multi-Scale Multi-View Detection (MSMVD).
MSMVD generates multi-scale BEV features by projecting multi-scale image features extracted from individual views into the BEV space, scale-by-scale.
Each of these BEV features inherits the properties of its corresponding scale image features from multiple views.
Therefore, these BEV features help the precise detection of pedestrians with consistently small or large scales in views.
Then, MSMVD combines information at different scales of multiple views by processing the multi-scale BEV features using a feature pyramid network.
This improves the detection of pedestrians with vastly different scales between views.
Extensive experiments demonstrate that exploiting multi-scale image features via multi-scale BEV features greatly improves the detection performance, and MSMVD outperforms the previous highest MODA by $4.5$ points on the GMVD dataset.

\end{abstract}

\section{Introduction}
\label{sec:intro}

Pedestrian detection from images is a crucial component in various applications such as surveillance systems~\cite{elhoseny2020multi}, sports analyses~\cite{cui2023sportsmot}, and human-computer interactions~\cite{wengefeld2019multi}.
In pedestrian detection, occlusion is a serious problem because it hinders detecting pedestrians hidden behind obstacles or other pedestrians~\cite{fleuret2007multicamera,chavdarova2018wildtrack}.
As a potential solution, many studies have focused on \textit{Multi-View Pedestrian Detection}~(MVPD)~\cite{chavdarova2018wildtrack,hou2020multiview,vora2023bringing}.
This task aims to detect pedestrians in the form of a bird's eye view (BEV) from multiple camera views.
In contrast to traditional monocular pedestrian detection~\cite{dollar2011pedestrian,zhang2017citypersons}, MVPD utilizes multiple calibrated cameras with partially overlapping fields of view.
MVPD is therefore expected to be more robust to occlusion in the overlapping areas than monocular pedestrian detection by aggregating complementary information across multiple views.

\begin{figure}[t]
    \centering
    \includegraphics[width=125mm]{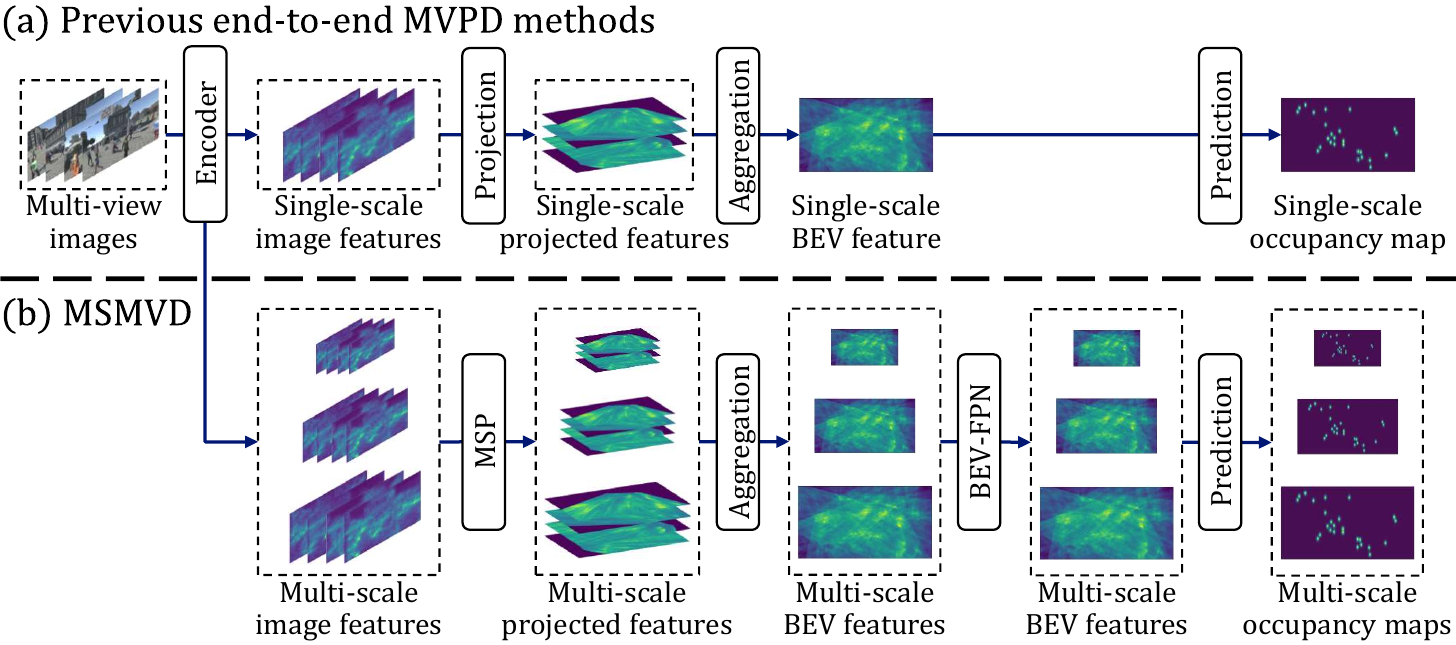}
    \caption{
    (a) Previous end-to-end MVPD methods generate a single-scale BEV feature from single-scale image features and predict one occupancy map. (b) MSMVD generates multi-scale BEV features from multi-scale image features and predicts multiple occupancy maps. In MSMVD, the image encoder consists of the backbone and image-FPN.
    }
    \label{fig:method}
\end{figure}

As with various computer vision tasks, end-to-end trainable deep learning methods are the dominant approach in MVPD~\cite{hou2020multiview,song2021stacked,hou2021multiview,vora2023bringing,engilberge2023two,qiu20223d,aung2024enhancing,suzuki2024scene,aung2024mvpocc,zhang2024mahalanobis,hwang2024booster}.
They adopt a unified framework, as shown in Fig.~\ref{fig:method}(a).
It first extracts image features for individual views from a specific layer of the image encoder (\eg, the last convolutional layer in ResNet~\cite{he2016deep}).
Then, it generates a BEV feature by projecting the image features to the BEV space based on calibrated camera parameters and predicts a BEV occupancy map from that BEV feature.
This framework aggregates complementary information from multiple views through the BEV feature, enabling the model to train in an end-to-end manner.

\begin{figure}[t]
    \centering
    \includegraphics[width=125mm]{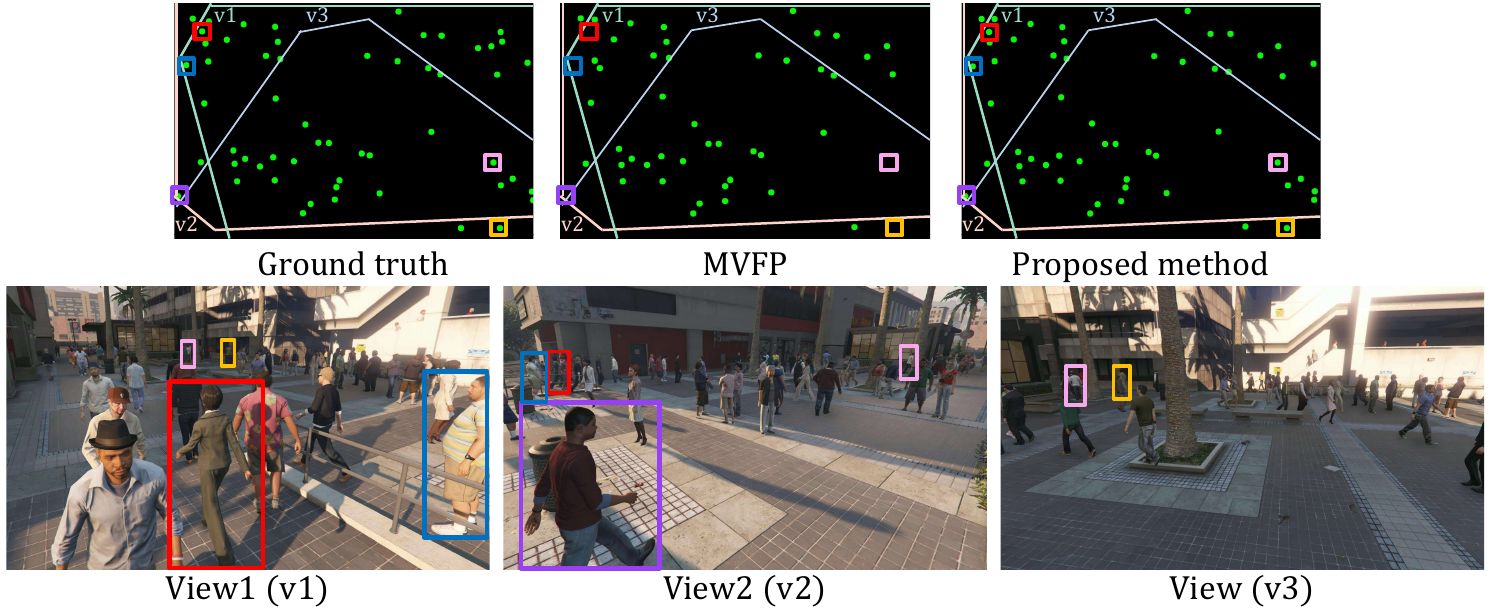}
    \caption{
    (Top) Comparison of the predicted BEV maps using MVFP~\cite{aung2024enhancing} and our proposed method. MVFP corresponds to Fig.~\ref{fig:method}(a). Green circles represent detected pedestrian locations, and colored rectangles illustrate the differences between MVFP and our proposed method. (Bottom) View images. Pedestrians with rectangles correspond to those with the same color rectangles on the BEV maps.
    }
    \label{fig:result}
\end{figure}

End-to-end MVPD methods have outperformed non-end-to-end methods~\cite{fleuret2007multicamera,roig2011conditional,xu2016multi,baque2017deep,chavdarova2017deep,chavdarova2018wildtrack}.
However, as shown in the result of MVFP~\cite{aung2024enhancing} in Fig.~\ref{fig:result}, they often struggle to detect two types of pedestrians: those with consistently small or large scales (i.e., apparent sizes in images) in views (yellow, pink, and purple rectangles), and those with vastly different scales between views (red and blue rectangles).
This is because single-scale image features from a specific layer of the image encoder used in existing end-to-end methods to generate the BEV feature have limited ability to represent pedestrians with diverse scales.
In contrast to end-to-end MVPD methods, many monocular methods~\cite{lin2017feature,ge2021yolox,chen2025yolo,khan2022f2dnet,khan2023localized} utilize multi-scale image features to effectively handle pedestrians with various scales, which are multiple features at different resolutions extracted from different layers in the image encoder.
Small pedestrians are better represented in high-resolution features and vice versa, so conducting prediction for each of the multi-scale image features helps the detection of pedestrians across a wide range of scales.
Despite their benefits, multi-scale image features have not been used in end-to-end MVPD methods.
Furthermore, in MVPD, since pedestrians often have vastly different scales between views, the important scale information of each view for detecting such pedestrians varies from one view to another.
Therefore, to fully utilize multi-scale image features in MVPD, it is important to combine information at different scales of multiple views.

In this paper, we propose a novel end-to-end MVPD method, called \textbf{M}ulti-\textbf{S}cale \textbf{M}ulti-\textbf{V}iew \textbf{D}etection (MSMVD), as shown in Fig.~\ref{fig:method}(b).
The main idea of MSMVD is to effectively generate and leverage multi-scale BEV features that inherit the properties of multi-scale image features from multiple views.
To this end, MSMVD newly introduces two key components: a multi-scale projection (MSP) and a BEV feature pyramid network (BEV-FPN).
Unlike existing methods that project only single-scale image features to the BEV space, MSP projects the multi-scale image features extracted by the image encoder to that space, scale-by-scale, as shown in Fig.~\ref{fig:component}(a).
This design allows MSMVD to generate multiple BEV features at different resolutions (i.e., multi-scale BEV features), each of which inherits the properties of the image features at its corresponding scale from multiple views.
Therefore, utilizing these multi-scale BEV features leads to precisely detecting pedestrians with consistently small or large scales in views.
BEV-FPN combines information at different scales of multiple views by processing the multi-scale BEV features utilizing an FPN-like network.
This allows MSMVD to effectively detect pedestrians with vastly different scales across views by combining important scale information of such pedestrians across multiple views. 
Extensive experiments demonstrate that exploiting multi-scale image features of multiple views via multi-scale BEV features leads to better detection performances.
Particularly in terms of MODA metric~\cite{chavdarova2018wildtrack}, MSMVD outperforms previous state-of-the-art methods by $4.5$ points on GMVD~\cite{vora2023bringing} and $0.5$ points on Wildtrack~\cite{chavdarova2018wildtrack} and MultiviewX~\cite{hou2020multiview}.
%

\section{Related Work}
\label{sec:related}

\paragraph{Monocular pedestrian detection.}
One of the main difficulties in monocular pedestrian detection is effectively handling pedestrians at various scales.
Traditional methods tackled this difficulty by generating hand-crafted features such as HOG features~\cite{dalal2005histograms} or SIFT features~\cite{lowe2004distinctive} for multiple images with different resolutions.
Modern deep learning methods~\cite{liu2016ssd,lin2017feature,liu2018path,tan2020efficientdet,zhu2021deformable,zhao2024detrs,liu2023center,khan2022f2dnet,khan2023localized} usually exploit multi-scale image features extracted from different layers in the backbone (\eg, ResNet~\cite{he2016deep}), where small pedestrians are better represented and detected in high-resolution features and vice versa.
As a pioneering work, SSD~\cite{liu2016ssd} leverages multi-scale image features and demonstrates their effectiveness.
FPN~\cite{lin2017feature} enhances multi-scale image features using a top-down path that gradually fuses low-resolution image features with higher-resolution image features.
PAFPN~\cite{liu2018path,ge2021yolox,lyu2022rtmdet,chen2025yolo} further improved FPN by adding an extra bottom-up path that gradually fuses output multi-scale features of FPN from high to low resolution.
EfficientDet~\cite{tan2020efficientdet} improves the detection performance by repeating a simplified version of PAFPN several times.
Deformable DETR~\cite{zhu2021deformable} and RT-DETR~\cite{zhao2024detrs} utilize Transformer~\cite{vaswani2017attention} to generate effective multi-scale image features.
CSP~\cite{liu2023center}, F2DNet~\cite{khan2022f2dnet}, and LSFM~\cite{khan2023localized} generate a powerful single-scale image feature by merging multi-scale image features of FPN into one feature.
Due to its simplicity yet effectiveness, PAFPN is utilized by our method for the image encoder and BEV-FPN.

\paragraph{Multi-view pedestrian detection.}
In MVPD, aggregating information on multiple views is essential to reduce the effect of occlusion because each view has an overlapping but different field of view~\cite{chavdarova2018wildtrack,hou2020multiview,vora2023bringing}.
Early methods~\cite{fleuret2007multicamera,roig2011conditional,xu2016multi,baque2017deep,chavdarova2017deep,chavdarova2018wildtrack} detected pedestrians for each view using monocular pedestrian detection and aggregated the detection results of multiple views.
While they were important early attempts in MVPD, they still relied on monocular detection, so the occlusion problem remained.
To overcome this problem, end-to-end trainable deep learning methods that do not rely on monocular detection have become the dominant approach~\cite{hou2020multiview,song2021stacked,hou2021multiview,vora2023bringing,engilberge2023two,qiu20223d,aung2024enhancing,suzuki2024scene,aung2024mvpocc,zhang2024mahalanobis,hwang2024booster}.
They generate a representation of the BEV space, called the BEV feature, by projecting image features of multiple views to the BEV space and predict a BEV occupancy map from the BEV feature.
The information on multiple views is aggregated through this BEV feature.
MVDet~\cite{hou2020multiview} is a pioneering end-to-end method that generates the BEV feature using inverse perspective mapping.
To further improve detection performance, recent studies have proposed more effective projections~\cite{hou2021multiview,song2021stacked,qiu20223d,hwang2024booster,aung2024enhancing}, data augmentations~\cite{hou2021multiview,engilberge2023two,qiu20223d,suzuki2024scene}, network architectures~\cite{engilberge2023two,aung2024enhancing,aung2024mvpocc}, and training strategies~\cite{vora2023bringing,zhang2024mahalanobis}.
While they have improved the detection performances, exploring the effective way to exploit multi-scale image features of multiple views remains an open problem.

\section{Proposed Method}
\label{sec:method}

\subsection{Overall Architecture}
\label{sub:arch}

Figure~\ref{fig:method}(b) shows an overview of MSMVD.
This takes $N$ view images $\{ I^{n} \}_{n=1}^{N}$, where $I^{n} \in \mathbb{R}^{3 \times H \times W}$ is the image from the $n$-th view.
Here, $H$ and $W$ are the height and width of the image, respectively.
The image encoder, consisting of the backbone and image feature pyramid network (image-FPN), extracts multi-scale image features for individual views.
Specifically, when using the ResNet~\cite{he2016deep} backbone, it first extracts image features $\{ F_{3}^{n}, F_{4}^{n}, F_{5}^{n}\}_{n=1}^{N}$, where $F_{l}^{n} \in \mathbb{R}^{C_{l} \times \frac{H}{2^{l}} \times \frac{W}{2^{l}}}$ is the $n$-th view image feature from the $l$-th stage in ResNet, and $C_{l}$ is the channel size in the $l$-th stage.
Then, as in many monocular methods~\cite{ge2021yolox,lyu2022rtmdet,chen2025yolo}, image-FPN enhances these multi-scale image features, resulting in $\{ \tilde{F}_{3}^{n}, \tilde{F}_{4}^{n}, \tilde{F}_{5}^{n} \}_{n=1}^{N}$, where $\tilde{F}_{l}^{n} \in \mathbb{R}^{C\times \frac{H}{2^{l}} \times \frac{W}{2^{l}}}$ is the enhanced $n$-th view image feature, and $C$ is the channel size of the image-FPN.
For these multi-scale image features, small pedestrians are better represented in $\tilde{F}_{3}^{n}$, and large pedestrians are better represented in $\tilde{F}_{5}^{n}$.

After obtaining multi-scale image features, our multi-scale projection (MSP) and subsequent max pooling generate multi-scale BEV features from those image features, and then, BEV-FPN combines information at different scales of multiple views.
MSP projects multi-scale image features $\{ \tilde{F}_{3}^{n}, \tilde{F}_{4}^{n}, \tilde{F}_{5}^{n} \}_{n=1}^{N}$ to the BEV space based on camera parameters, scale-by-scale. 
We denote these projected features as $\{ P_{3}^{n}, P_{4}^{n}, P_{5}^{n} \}_{n=1}^{N}$, where $P_{l}^{n} \in \mathbb{R}^{C \times \frac{X}{2^{l-2}} \times \frac{Y}{2^{l-2}}}$ is obtained from $\tilde{F}_{l}^{n}$.
Here, $X$ and $Y$ are the width and height of the ground truth BEV map.
This MSP allows the model to generate multi-scale BEV features that inherit the properties of multi-scale image features of multiple views.
We formulate MSP in Sec.~\ref{sub:proj}.
To aggregate the information at each scale of multiple views, we perform max pooling along the view direction, scale-by-scale, and generate multi-scale BEV features $\{ B_{3}, B_{4}, B_{5} \}$.
Here, $B_{l} \in \mathbb{R}^{C \times \frac{X}{2^{l-2}} \times \frac{Y}{2^{l-2}}}$ is generated from $\{ P_{l}^{n} \}_{n=1}^{N}$.
Namely, the information on small pedestrians of multiple views is aggregated to $B_{3}$, and that on large pedestrians is aggregated to $B_{5}$.
BEV-FPN combines the information at different scales of multiple views by processing multi-scale BEV features, resulting in refined multi-scale BEV features $\{ \tilde{B}_{3}, \tilde{B}_{4}, \tilde{B}_{5} \}$, where $\tilde{B}_{l} \in \mathbb{R}^{C \times \frac{X}{2^{l-2}} \times \frac{Y}{2^{l-2}}}$.
We describe the detailed structure of the BEV-FPN in Sec.~\ref{sub:fpn}.

Then, prediction networks followed by the sigmoid function predict multi-scale BEV occupancy maps $\{ M_{3}, M_{4}, M_{5} \}$ from the multi-scale BEV features, where $M_{l} \in \mathbb{R}^{1 \times \frac{X}{2^{l-2}} \times \frac{Y}{2^{l-2}}}$ is predicted from $\tilde{B}_{l}$.
Because the predicted occupancy maps have lower resolutions than the ground truth, we also predict offset maps $\{ O_{3}, O_{4}, O_{5} \}$ to recover the discretization errors of pedestrian locations, following MVDeTr~\cite{hou2021multiview}.
Here, $O_{l} \in \mathbb{R}^{2 \times \frac{X}{2^{l-2}} \times \frac{Y}{2^{l-2}}}$ is predicted from $\tilde{B}_{l}$.
MSMVD is optimized by calculating the loss for all occupancy maps and offset maps during training and predicts the final occupancy map by merging multi-scale occupancy maps during inference.
We describe these training and inference procedures in Sec.~\ref{sub:train}.

\subsection{Multi-scale Projection}
\label{sub:proj}

\begin{figure}[tb]
    \centering
    \includegraphics[width=125mm]{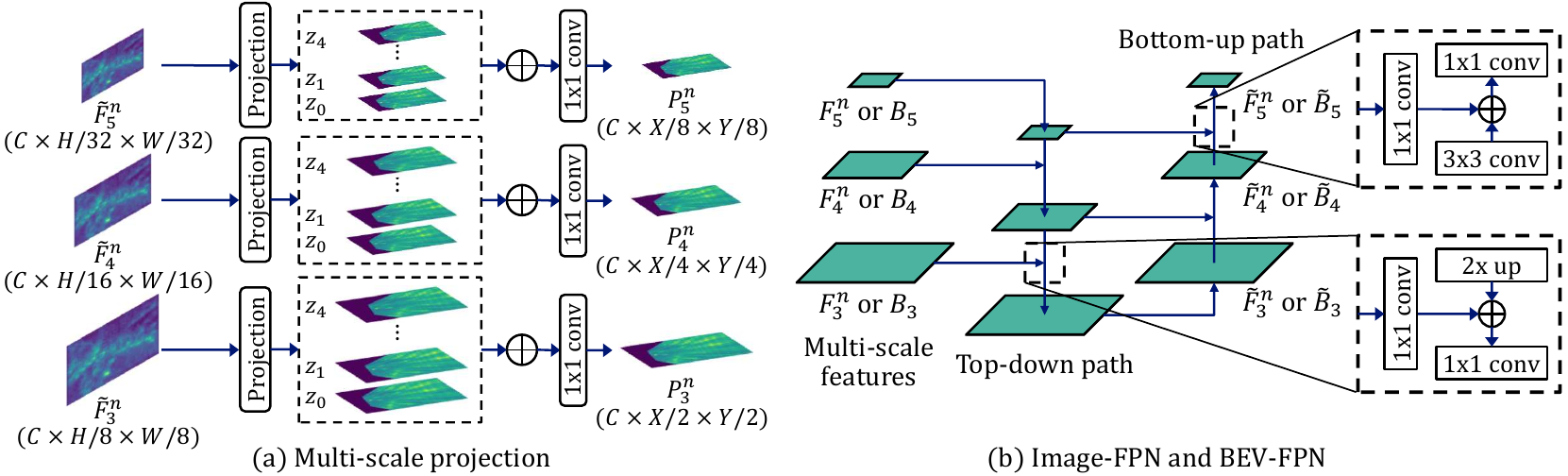}
    \caption{
    (a) Multi-scale projection (MSP). (b) Image-FPN and BEV-FPN consist of top-down and bottom-up paths. $\oplus$ indicates a concatenation. Image-FPN processes multi-scale image features, and BEV-FPN processes multi-scale BEV features.
    }
    \label{fig:component}
\end{figure}

Figure~\ref{fig:component}(a) shows our multi-scale projection (MSP).
To inherit the properties of multi-scale image features of multiple views to the BEV feature, MSP projects multi-scale image features $\{ \tilde{F}_{3}^{n}, \tilde{F}_{4}^{n}, \tilde{F}_{5}^{n} \}_{n=1}^{N}$ to the BEV space and output projected features $\{ P_{3}^{n}, P_{4}^{n}, P_{5}^{n} \}_{n=1}^{N}$.
Let $K^{n} \in \mathbb{R}^{3 \times 3}$ and $\lbrack R^{n} | T^{n} \rbrack \in \mathbb{R}^{3 \times 4}$ be the intrinsic and extrinsic camera parameter matrices of the $n$-th view, respectively.
Here, $R^{n}$ represents the rotation, and $T^{n}$ represents the translation.
We project $n$-th view multi-scale image features to the BEV space using the correspondence~\cite{hou2020multiview,hou2021multiview,song2021stacked,qiu20223d} between the $n$-th view 2D image pixel coordinates $(u^{n}, v^{n})$ and the 3D world position $(x, y, z)$, as
\begin{align}
    \centering
    \gamma^{n} \begin{pmatrix}
        u^{n} \\
        v^{n} \\
        1
    \end{pmatrix} = K^{n} \lbrack R^{n} | T^{n} \rbrack \begin{pmatrix}
        x \\
        y \\
        z \\
        1
    \end{pmatrix},
\label{formula:proj}
\end{align}
where $\gamma^{n}$ is a scaling factor that accounts for a down-sampling of image features and determines the spatial resolution of the BEV feature.
For each of $n$-th view multi-scale image features, we set $\gamma^{n}$ to the value that makes the resolution ratio of projected features $\{ P_{3}^{n}, P_{4}^{n}, P_{5}^{n} \}$ and that of the image features $\{ \tilde{F}_{3}^{n}, \tilde{F}_{4}^{n}, \tilde{F}_{5}^{n} \}$ the same.
Specifically, for $\tilde{F}_{l}^{n}$, $\gamma^{n}$ is set so that the spatial resolution of $P_{l}^{n}$ is $\frac{X}{2^{l-2}} \times \frac{Y}{2^{l-2}}$.
This allows the model to inherit the properties of multi-scale image features to multi-scale BEV features.
We project each image feature $\tilde{F}_{l}^{n}$ with multiple heights $\{ z_{0}, z_{1}, z_{2}, z_{3}, z_{4} \}$, like 3DROM~\cite{qiu20223d}, to utilize information about the entire pedestrian body, where $z_{i}$ corresponds to the height of $30 \times i $ cm.
Then, projected features of multiple heights are concatenated along the channel direction and processed by a $1 \times 1$ convolutional layer, resulting in a final $C$-channel projected feature $P_{l}^{n}$.

\subsection{BEV-FPN}
\label{sub:fpn}

BEV-FPN combines information at different scales of multiple views to precisely detect pedestrians whose scales are vastly different between views because important scale features of such pedestrians differ from one view to another.
To this end, we utilize PAFPN~\cite{liu2018path} for BEV-FPN, which consists of the top-down and bottom-up paths, as shown in Fig.~\ref{fig:component}(b).
The top-down path takes multiple features at different resolutions as input and gradually fuses low-resolution features with higher-resolution features.
This process combines important information in low-resolution features with that in higher-resolution features.
The bottom-up path takes the multiple features output from the top-down path as input and gradually fuses high-resolution features with lower-resolution features.
This process combines important information in high-resolution features with that in lower-resolution features.
Since the information of $\{ P_{l}^{n} \}_{n=1}^{N}$ is aggregated to $B_{l}$ by MSP and max pooling, BEV-FPN combines information at different scales of multiple views by processing multi-scale BEV features $\{ B_{3}, B_{4}, B_{5} \}$.
As a result, BEV-FPN outputs enhanced multi-scale BEV features $\{ \tilde{B}_{3}, \tilde{B}_{4}, \tilde{B}_{5} \}$, and they improve the detection of pedestrians with vastly different scales between views.

\subsection{Training and Inference}
\label{sub:train}

During training, we calculate the loss for multi-scale BEV occupancy maps $\{ M_{3}, M_{4}, M_{5}\}$ and multi-scale offset maps $\{ O_{3}, O_{4}, O_{5}\}$.
Following MVDeTr~\cite{hou2021multiview}, we used the Focal loss~\cite{lin2017focal} and L1 loss variant~\cite{zhou2019objects,liu2023center} for the occupancy maps and offset maps, respectively.
Let $\mathcal{L}_\mathrm{det} (\cdot, \cdot)$ be the loss function for the occupancy maps, and $\mathcal{L}_\mathrm{off} (\cdot, \cdot)$ be the loss function for the offset maps.
Our overall loss $\mathcal{L}_\mathrm{all}$ is 
\begin{align}
\centering
    &\mathcal{L}_\mathrm{all} = \sum_{l \in \{ 3, 4, 5 \}} \{ \mathcal{L}_\mathrm{det} (M_{l}, \hat{M}_l) +  \mathcal{L}_\mathrm{off} (O_{l}, \hat{O}_l) \}.
\label{formula:loss_i}
\end{align}
Here, $\hat{M}_{l}$ is the ground truth occupancy map down-sampled to match with $M_{l}$ and smoothed using a Gaussian kernel.
$\hat{O}_{l}$ is the ground truth offset map corresponding to the discretization error of $\hat{M}_{l}$.
While we do not use $O_{4}$ and $O_{5}$ for inference, we calculate the losses for them as auxiliary losses.
We optimize MSMVD using $\mathcal{L}_\mathrm{all}$ as the overall training objective.

During inference, we merge predicted multiple occupancy maps $\{ M_{3}, M_{4}, M_{5} \}$ into one occupancy map $M \in \mathbb{R}^{1 \times \frac{X}{2} \times \frac{Y}{2}}$ to combine pedestrians with various scales, as
\begin{align}
\centering
    &M = \frac{1}{3} \{ M_{3} + \mathrm{Upsample}(M_{4}) + \mathrm{Upsample}(M_{5}) \},
\label{formula:inference}
\end{align}
where the $\mathrm{Upsample}$ function indicates up-sampling of the occupancy map to the same spatial resolution as $M_{3}$ using a bilinear interpolation.
Since down-sampling high-resolution occupancy maps could lose detailed location information of pedestrians, we up-sample low-resolution occupancy maps to the highest resolution.
After the merging, we extract pedestrians over the threshold and convert their locations to those in the original $X \times Y$ BEV map using the offset map $O_{3}$ in the same way as MVDeTr~\cite{hou2021multiview}.

\section{Experiments}
\label{sec:exp}

\subsection{Datasets}
\label{sub:data}

\noindent
\textbf{Wildtrack}~\cite{chavdarova2018wildtrack}.
This is a real-world dataset consisting of images captured with $7$ cameras and comprises $400$ frames.
This dataset covers a $12 \, \mathrm{m} \times 36 \, \mathrm{m}$ region quantized into a $480 \times 1440$ grid using square grid cells of $2.5 \, \mathrm{cm^2}$.
Each frame includes $20$ pedestrians on average.
Wildtrack splits them into $360$ frames for training and the remaining $40$ frames for testing.
We use the randomly sampled $40$ frames in the training split as the validation data.

\noindent
\textbf{MultiviewX}~\cite{hou2020multiview}.
This is a synthetic dataset using the Unity engine and closely follows the style of Wildtrack.
This dataset comprises $400$ frames captured by $6$ cameras and covers a $16 \, \mathrm{m} \times 25 \, \mathrm{m}$ region.
The region is quantized into a $640 \times 1000$ grid.
Each frame includes $40$ pedestrians on average.
MultiviewX also splits them into $360$ frames for training and $40$ frames for testing.
We use $40$ frames in the training split as the validation data.

\noindent
\textbf{GMVD}~\cite{vora2023bringing}.
This is a large-scale synthetic dataset and includes $7$ scenes.
Each scene varies the number of cameras and camera layouts, which makes the dataset more challenging than a constrained camera setup like Wildtrack or MultiviewX.
Except for the camera setup and the size of the covered region, all parameters follow those of MultiviewX.
Each frame includes $20$-$40$ pedestrians on average.
GMVD splits them into $6$ scenes with $4983$ frames for training and $1$ scene with $1012$ frames for testing.
Following Vora~\etal~\cite{vora2023bringing}, we use MultiviewX as the validation data during training on the GMVD train split and evaluate trained models on the GMVD test split.
Since GMVD is the largest dataset, we set it as the main dataset for our experiments.

\subsection{Implementation Details and Evaluation Metrics}
\label{sub:imple}

\begin{table}[t]
\centering
\scalebox{0.85}{
\begin{tabular}{l|cccc} \hline
Method & MODA & MODP & Precision & Recall \\ \hline
$\textrm{Vora}+$~\cite{vora2023bringing} & 68.2 & 76.3 & 91.5 & 75.5 \\ 
$\textrm{SHOT}^{\dagger}$~\cite{song2021stacked} & 71.5 & 77.9 & 93.7 & 76.7 \\
$\textrm{Suzuki}+$~\cite{suzuki2024scene} & 72.3 & 77.1 & 91.3 & 77.1 \\ 
$\textrm{3DROM}^{\dagger}$~\cite{qiu20223d} & 73.7 & 77.3 & 92.2 & 80.5 \\
$\textrm{MVAug}^{\dagger}$~\cite{engilberge2023two} & 73.8 & 77.9 & 91.9 & 80.8 \\
$\textrm{OmniOcc}^{\dagger}$~\cite{aung2024mvpocc} & 75.1 & 76.9 & 92.3 & 82.0 \\
$\textrm{MVFP}^{\dagger}$~\cite{aung2024enhancing} & 75.7 & 78.2 & 94.3 & 80.5 \\
Ours & \textbf{80.2} & \textbf{81.3} & \textbf{95.7} & \textbf{83.9} \\ \hline
\end{tabular}
}
\caption{Comparison on GMVD. $\dagger$ indicates our re-implementation results.}
\label{table:gmvd}
\end{table}

\begin{table}[t]
\centering
\scalebox{0.85}{
\begin{tabular}{l|cccc|cccc} \hline
\multirow{2}{*}{Method} & \multicolumn{4}{|c}{Wildtrack} & \multicolumn{4}{|c}{MultiviewX} \\
& MODA & MODP & Precision & Recall & MODA & MODP & Precision & Recall \\ \hline
$\textrm{Vora}+$~\cite{vora2023bringing} & 86.7 & 76.2 & 95.1 & 91.4 & 88.2 & 79.9 & 96.8 & 91.2 \\ 
MVDet~\cite{hou2020multiview} & 88.2 & 75.7 & 94.7 & 93.6 & 83.9 & 79.6 & 96.8 & 86.7\\ 
SHOT~\cite{song2021stacked} & 90.2 & 76.5 & 96.1 & 94.0 & 88.3 & 82.0 & 96.6 & 91.5 \\
MVDeTr~\cite{hou2021multiview} & 91.5 & 82.1 & \textbf{97.4} & 94.0 & 93.7 & \textbf{91.3} & 99.5 & 94.2 \\
M-MVOT~\cite{zhang2024mahalanobis} & 92.1 & 81.3 & 94.5 & 97.8 & 96.7 & 86.1 & 98.8 & \textbf{97.9} \\
MVAug~\cite{engilberge2023two} & 93.2 & 79.8 & 96.3 & 97.0 & 95.3 & 89.7 & 99.4 & 95.9 \\
3DROM~\cite{qiu20223d} & 93.5 & 75.9 & 97.2 & 96.2 & 95.0 & 84.9 & 99.0 & 96.1 \\
OmniOcc~\cite{aung2024mvpocc} & 93.5 & 81.5 & 94.9 & 97.8 & 94.4 & 88.7 & 98.7 & 95.6 \\
MVFP~\cite{aung2024enhancing} & 94.1 & 78.8 & 96.4 & 97.7 & 95.7 & 85.1 & 98.4 & 97.2 \\
Ours & \textbf{94.6} & \textbf{83.3} & 95.9 & \textbf{98.8} & \textbf{97.2} & \textbf{91.3} & \textbf{99.8} & 97.4 \\ \hline
\end{tabular}
}
\caption{Comparison on Wildtrack and MultiviewX. OmniOcc is re-implemented by us.}
\label{table:wildtrack_multiviewx}
\end{table}

Unless otherwise mentioned, we employed ResNet18~\cite{he2016deep} and PAFPN~\cite{liu2018path} for the backbone and image-FPN in the image encoder, respectively.
ResNet18 was pretrained on ImageNet~\cite{deng2009imagenet}.
We use four convolutional layers for each prediction network.
We set the channel size $C$ to $256$ and the detection threshold to $0.4$.

Input images were resized to $720 \times 1280$.
Following MVDeTr~\cite{hou2021multiview}, we applied random resizing and cropping to images as the data augmentation during training.
We optimized the model using the Adam optimizer~\cite{loshchilov2018decoupled}.
We set the batch size to $1$ and accumulated the gradient over $16$ batches.
The learning rate was initialized to $1.0 \times 10^{-3}$ and decayed to $1.0 \times 10^{-6}$ following a cosine schedule.
We set the training epoch to $10$ for GMVD and $50$ for Wildtrack and MultiviewX.
More training details are provided in Supp.~\ref{supp:imple}.

Following previous studies~\cite{hou2020multiview,vora2023bringing}, we used four standard metrics provided by Chavdarova~\etal~\cite{chavdarova2018wildtrack} and Kasturi~\etal~\cite{kasturi2008framework}: multiple object detection accuracy (MODA), multiple object detection precision (MODP), recall, and precision.
A detected pedestrian was classified as a true positive if its distance from the ground truth was within $0.5$ meters.
We used MODA as the primary performance indicator following previous studies~\cite{hou2020multiview,vora2023bringing}.

\subsection{Comparison with Previous State-of-the-Art Methods}
\label{sub:sota}

To verify the effectiveness of MSMVD, we compared it with previous state-of-the-art methods on GMVD, as shown in Table~\ref{table:gmvd}.
Previous methods generate a single-scale BEV feature from single-scale image features, as shown in Fig.~\ref{fig:method}(a).
Since several previous methods had been evaluated only on Wildtrack and MultiviewX, we re-implemented them for GMVD.
Results of other methods are directly copied from the original papers.
MSMVD significantly outperformed the previous methods on all metrics.
In particular, it achieved MODA and MODP $4.5$ and $3.1$ points higher than the previous highest scores, respectively.
These results demonstrate that exploiting and combining multi-scale image features of multiple views leads to a better detection performance and show that MSMVD is superior to the other methods.
Figure~\ref{fig:result} visually compares the results of MSMVD and MVFP~\cite{aung2024enhancing}.
While MVFP failed to detect pedestrians with consistently small (yellow and pink rectangles) or large (a purple rectangle) scales in views and pedestrians with vastly different scales between views (red and blue rectangles), our MSMVD accurately detected these pedestrians.
This indicates the advantages of MSMVD over MVFP.
Specifically, the detection of the former pedestrians indicates the advantage of exploiting multi-scale image features and conducting prediction for each of the multi-scale BEV features, and that of the latter pedestrians indicates the advantage of combining the information at different scales of multiple views by BEV-FPN.

\begin{table}
\centering
\begin{tabular}{cc}
\bmvaHangBox{
\scalebox{0.69}{
\begin{tabular}{l|cccc} \hline
& MODA & MODP & Precision & Recall \\ \hline
Baseline    & 71.8 & 80.0 & 93.6 & 76.9 \\ 
+ MSP       & 74.9 & 80.5 & 94.3 & 80.3 \\
+ MSP and BEV-FPN   & \textbf{80.2} & \textbf{81.3} & \textbf{95.7} & \textbf{83.9}  \\ \hline
\end{tabular}
}}&
\bmvaHangBox{
\scalebox{0.69}{
\begin{tabular}{c|cccc} \hline
& MODA & MODP & Precision & Recall \\ \hline
$M_{3} \& O_{3}$ & 79.1 & 81.2 & 94.9 & 83.6 \\
$M_{4} \& O_{4}$ & 78.1 & 80.3 & 95.0 & 82.4 \\
$M_{5} \& O_{5}$ & 77.6 & 79.5 & 95.0 & 81.9 \\
Ours & \textbf{80.2} & \textbf{81.3} & \textbf{95.7} & \textbf{83.9} \\ \hline
\end{tabular}
}} \\
(a)&(b)
\end{tabular}
\caption{(a) Effect of each component in MSMVD. (b) Comparison of our multi-scale inference and single-scale inference.}
\label{tab:component}
\end{table}

\begin{table}[t]
\centering
\scalebox{0.85}{
\begin{tabular}{l|l|ccccc} \hline
Method & Backbone & MODA & MODP & Precision & Recall & Params (M) \\ \hline
\multirow{4}{*}{Baseline} & ResNet18 & 71.8 & 80.0 & 93.6 & 76.9 & 11.7 \\ 
& ResNet34 & 72.1 & 78.2 & 95.5 & 75.6 & 21.8 \\ 
& ResNet50 & 74.3 & 79.2 & 94.1 & 79.2 & 24.2 \\
& ResNet101 & 75.8 & 78.9 & 96.8 & 78.3 & 43.2 \\ \hline
\multirow{4}{*}{Ours} & ResNet18  & 80.2 & 81.3 & 95.7 & 83.9 & 22.9 \\ 
& ResNet34 & 81.0 & 81.5 & 94.8 & \textbf{85.7} & 33.0 \\ 
& ResNet50 & 82.0 & 81.3 & 94.9 & 85.6 & 35.9 \\
& ResNet101 & \textbf{82.2} & \textbf{82.6} & \textbf{97.2} & 85.6 & 54.9 \\ \hline
\end{tabular}
}
\caption{Effect of backbone size on the baseline model and our proposed method.}
\label{table:backbone}
\end{table}

To validate the versatility of MSMVD, which is not limited to GMVD, we also compared it with previous methods on Wildtrack and MultiviewX, as shown in Table~\ref{table:wildtrack_multiviewx}.
The result of OmniOcc~\cite{aung2024mvpocc} is achieved by our re-implementation, and the results of the other methods are directly copied from the original papers.
MSMVD outperformed the previous methods on almost all metrics.
In particular, it exceeded the previous highest MODA by $0.5$ points on both datasets.
These results demonstrate that our method is effective for various datasets.

\subsection{Ablation Study}
\label{sub:ablation}

In this subsection, we investigate the effect of MSP, BEV-FPN, our inference procedure, and the backbone size.
We set the model that removes MSP and BEV-FPN from MSMVD and generates a single-scale BEV feature from single-scale image features extracted by the last layer of the image encoder as the baseline.
All experiments were conducted on GMVD.
In Supp.~\ref{supp:ablation}, we also investigate the effect of the training with $O_{4}$ and $O_{5}$, the offset prediction, the pooling method, the architecture of image-FPN and BEV-FPN, and the scaling factor $\gamma^{n}$ in MSP.

\noindent
\textbf{Effect of each component.}
To investigate the contribution of MSP and BEV-FPN, we gradually added them to the baseline, as shown in Table~\ref{tab:component}(a).
Adding MSP significantly improved MODA and recall and outperformed all methods in Table~\ref{table:gmvd} on MODA, other than OmniOcc and MVFP.
This indicates that adding MSP reduces missed detection and demonstrates the effectiveness of utilizing multi-scale image features via multi-scale BEV features.
Adding BEV-FPN further improved the detection performance and outperformed all methods in Table~\ref{table:gmvd}.
This shows that adding BEV-FPN enhances multi-scale BEV features and demonstrates the importance of combining information at different scales of multiple views.

\noindent
\textbf{Effect of merging multi-scale occupancy maps.}
To investigate the effect of merging predictions at multiple scales during inference, we compared our inference procedure with using only a single-scale occupancy map and the corresponding offset map, as shown in Table~\ref{tab:component}(b).
When employing our inference procedure, we achieved a better detection result than when using only a single-scale occupancy map and its offset map.
This demonstrates the advantage of combining predictions at multiple scales.
In addition, even when using only a single-scale occupancy map and offset map, MSMVD outperformed all methods in Table~\ref{table:gmvd}.
This also indicates that MSMVD's BEV feature is superior to that of other methods.

\noindent
\textbf{Effect of the backbone size.}
MSMVD has more parameters than the baseline due to MSP and BEV-FPN.
To show that the performance improvement by MSMVD is not simply due to this parameter increase, we applied backbones with different sizes and investigated their effect on the detection performance, as shown in Table~\ref{table:backbone}.
While larger backbones improved the detection performance of the baseline, MSMVD with ResNet18 achieved a better result than the baseline with the same or more parameters.
This demonstrates that MSMVD is superior to a simple parameter increase by larger backbones.
Larger backbones are also beneficial to our method, and MSMVD with the largest ResNet101 achieved $82.2$ MODA and $82.6$ MODP, which significantly outperformed existing methods in Table~\ref{table:gmvd}.

\section{Conclusion}
\label{sec:conclusion}

We proposed a novel end-to-end multi-view pedestrian detection (MVPD) method, called MSMVD.
In contrast to previous methods that rely only on single-scale image features, MSMVD exploits multi-scale image features via multi-scale bird's eye view (BEV) features to effectively detect pedestrians at various scales.
In addition, MSMVD combines information at different scales of multiple views by processing multi-scale BEV features using a feature pyramid network.
This makes the model more robust to pedestrians with vastly different scales between views.
Extensive experiments demonstrate that exploiting multi-scale image features via multi-scale BEV features improves the detection performance, and MSMVD achieved a new state-of-the-art result on three major MVPD datasets.
We hope that our work opens new possibilities in exploring end-to-end MVPD.

\bibliography{main}

\clearpage
\appendix
\section*{Supplementary Material}

\setcounter{page}{1}

\begin{table}[ht]
\centering
\scalebox{0.90}{
\begin{tabular}{c|cccc} \hline
 & MODA & MODP & Precision & Recall \\ \hline
Train w/o $O_{4}$ and $O_{5}$ & 79.6 & 81.0 & 95.4 & 83.6 \\
Train w/ $O_{4}$ and $O_{5}$ & \textbf{80.2} & \textbf{81.3} & \textbf{95.7} & \textbf{83.9} \\ \hline
\end{tabular}
}
\caption{Effect of training with $O_{4}$ and $O_{5}$.}
\label{table:o4_o5}
\end{table}

\begin{table}[ht]
\centering
\begin{tabular}{cc}
\bmvaHangBox{
\scalebox{0.76}{
\begin{tabular}{c|cccc} \hline
 & MODA & MODP & Precision & Recall \\ \hline
w/o offset & 79.2 & 80.6 & 95.0 & 83.2 \\
w/ offset  & \textbf{80.2} & \textbf{81.3} & \textbf{95.7} & \textbf{83.9} \\ \hline
\end{tabular}
}}&
\bmvaHangBox{
\scalebox{0.76}{
\begin{tabular}{c|cccc} \hline
Pooling & MODA & MODP & Precision & Recall \\ \hline
Mean & 79.7 & 81.2 & 94.9 & 83.6 \\
Max & \textbf{80.2} & \textbf{81.3} & \textbf{95.7} & \textbf{83.9} \\ \hline
\end{tabular}
}} \\
(a)&(b)
\end{tabular}
\caption{(a) Effect of the offset prediction. (b) Effect of the pooling method.}
\label{table:offset_pool}
\end{table}

\section{Training Details}
\label{supp:imple}

When generating the ground truth occupancy maps, we set the diameter of the Gaussian kernel to $20$, $10$, and $5$ pixels for $M_{3}$, $M_{4}$, and $M_{5}$, respectively.
We set the scale range of random resizing and cropping to $[0.8, 1.2]$ during training.
For the Adam optimizer~\cite{loshchilov2018decoupled}, we set the optimizer momentum to $\beta_1 = 0.9$ and $\beta_2 = 0.999$.
We did not use the weight decay.
All experiments were conducted on an Nvidia 80GB A100 GPU.

\begin{table}[t]
\centering
\scalebox{0.90}{
\begin{tabular}{cc|cccc} \hline
Image-FPN bottom-up & BEV-FPN bottom-up & MODA & MODP & Precision & Recall \\ \hline
           && 76.4 & 80.4 & 95.0 & 80.7 \\
\checkmark && 78.2 & \textbf{81.3} & 95.6 & 81.9 \\
           & \checkmark & 78.3 & 80.8 & \textbf{96.2} & 81.5 \\
\checkmark & \checkmark & \textbf{80.2} & \textbf{81.3} & 95.7 & \textbf{83.9} \\ \hline
\end{tabular}
}
\caption{Effect of the bottom-up path in image-FPN and BEV-FPN.}
\label{table:bottom_up}
\end{table}

\begin{table}[t]
\centering
\scalebox{0.90}{
\begin{tabular}{c|cccc} \hline
Resolution of $B_{3}$, $B_{4}$, and $B_{5}$ & MODA & MODP & Precision & Recall \\ \hline
Same & 78.4 & 80.6 & 94.5 & 83.3 \\
Different (Ours) & \textbf{80.2} & \textbf{81.3} & \textbf{95.7} & \textbf{83.9} \\ \hline
\end{tabular}
}
\caption{Effect of the scaling factor $\gamma^{n}$ in MSP.}
\label{table:scale}
\end{table}

\section{Additional Ablation Study}
\label{supp:ablation}

In this section, we investigate the effect of the training with $O_{4}$ and $O_{5}$, the offset prediction, the pooling method, the architecture of image-FPN and BEV-FPN, and the scaling factor $\gamma^{n}$ in MSP.
All experiments were conducted on GMVD.

\paragraph{Effect of training with $O_{4}$ and $O_{5}$.}
While we do not use offset maps $O_{4}$ and $O_{5}$ during inference, we calculate the losses for not only $O_{3}$ but also $O_{4}$ and $O_{5}$ during training.
To investigate this effect, we compared MSMVD trained with and without the losses for $O_{4}$ and $O_{5}$.
As shown in Table~\ref{table:o4_o5}, the model trained with the losses for $O_{4}$ and $O_{5}$ achieved a better detection result than one without them.
We presume that this result is because the losses for $O_{4}$ and $O_{5}$ worked as auxiliary losses to precisely predict occupancy maps at their corresponding scales, leading to more accurate occupancy map predictions.

\paragraph{Effect of offset prediction.}
Some previous end-to-end MVPD methods~\cite{hou2020multiview,song2021stacked,qiu20223d} do not predict the offset map, but only predict the occupancy map, ignoring the discretization errors of pedestrian locations.
To investigate the effect of this offset, we compared MSMVD with and without the offset prediction, as shown in Table~\ref{table:offset_pool}(a).
MSMVD with the offset prediction achieved a better detection result than without it.
This result demonstrates the importance of the offset prediction to localize pedestrians precisely.

\paragraph{Pooling method.}
We performed max pooling to aggregate projected features from multiple views, as described in Sec.~\ref{sub:arch}.
Another option is to perform mean pooling instead of max pooling.
When we compared these two pooling operations, as shown in Table~\ref{table:offset_pool}(b), performing max pooling achieved a better detection performance.
Max pooling operation extracts the most relevant and informative features from different view perspectives for the subsequent detection modules, which presumably leads to better detection performance.

\paragraph{Architecture of image-FPN and BEV-FPV.}
We exploited PAFPN~\cite{liu2018path,ge2021yolox,lyu2022rtmdet,chen2025yolo}, consisting of not only the top-down path but also the bottom-up path, for both image-FPN and BEV-FPN.
We compared MSMVD with and without this bottom-up path in image-FPN and BEV-FPN to investigate its effect, as shown in Table~\ref{table:bottom_up}.
Adding the bottom-up path to image-FPN improved the detection performance, particularly $1.8$ MODA.
This demonstrates that the bottom-up path in image-FPN enables the model to extract more effective multi-scale image features, resulting in better detection.
Adding the bottom-up path to BEV-FPN also improved the detection performance, particularly $1.9$ MODA.
This demonstrates that the bottom-up path in BEV-FPN enhances the combination of information at different scales of multiple views, resulting in more effective multi-scale BEV features.
Adding the bottom-up path to both image-FPN and BEV-FPN achieved the best detection result.
This indicates that the bottom-up path is important for both image-FPN and BEV-FPN to achieve better detection performance.

\paragraph{Effect of the scaling factor $\gamma^{n}$.}
In MSP, we set the scaling factor $\gamma^{n}$ so that the resolution ratio between multi-scale BEV features $\{ B_{3}, B_{4}, B_{5} \}$ and the resolution ratio between multi-scale image features $\{ \tilde{F}_{3}^{n}, \tilde{F}_{4}^{n}, \tilde{F}_{5}^{n} \}$ are the same, as described in Sec.~\ref{sub:proj}.
To investigate this effect, we compared our $\gamma^{n}$ setting with the case where $\gamma^{n}$ is set so that the resolutions of $B_{4}$ and $B_{5}$ were the same as $B_{3}$ (i.e., $\frac{X}{2} \times \frac{Y}{2}$), as shown in Table~\ref{table:scale}.
Adopting our $\gamma^{n}$ setting achieved a better detection performance.
This result demonstrates the importance of maintaining the resolution ratio to fully utilize the information at multiple scales of multiple views when generating multi-scale BEV features from multi-scale image features.

\end{document}